\pdfoutput=1

\documentclass[11pt]{article}

\usepackage[]{acl}

\usepackage{times}
\usepackage{latexsym}
\usepackage{soul}
\usepackage{booktabs}
\usepackage{graphicx}
\usepackage{multirow}
\usepackage{tikz}
\usepackage{amsmath}

\usepackage[T1]{fontenc}

\usepackage[utf8]{inputenc}

\usepackage{microtype}

\newcommand{\astfootnote}[1]{
    \let\oldthefootnote=\thefootnote
    \setcounter{footnote}{1}
    \renewcommand{\thefootnote}{\fnsymbol{footnote}}
    \footnotetext{#1}
    \let\thefootnote=\oldthefootnote
}

\newcommand{\newtextcircled}[1]{\raisebox{.5pt}{\textcircled{\raisebox{-.9pt} {#1}}}}

\title{HYU at SemEval-2022 Task 2: Effective Idiomaticity Detection with Consideration at Different Levels of Contextualization}

\author{Youngju Joung$^*$ \\
Dept. of Industrial Security \\ 
Dept. of Applied Statistics \\
Chung Ang University \\
\texttt{ojoo.yj@gmail.com} \\\And
Taeuk Kim$^{**}$ \\
Dept. of Computer Science \\
Dept. of Artificial Intelligence \\
Hanyang University \\
\texttt{kimtaeuk@hanyang.ac.kr} \\}

\begin{document}
\maketitle
\begin{abstract}
We propose a unified framework that enables us to consider various aspects of contextualization at different levels to better identify the idiomaticity of multi-word expressions.
Through extensive experiments, we demonstrate that our approach based on the inter- and inner-sentence context of a target MWE is effective in improving the performance of related models.
We also share our experience in detail on the task of SemEval-2022 Task 2 such that future work on the same task can be benefited from this.
\end{abstract}

\astfootnote{Work done while Youngju was an intern at HYU.}
\setcounter{footnote}{0}
\astfootnote{$^*$Corresponding author.}
\setcounter{footnote}{0}

\section{Introduction}

Multi-word expressions (MWEs) are a group of linguistic components containing two or more words with outstanding collocation \cite{baldwin2010multiword,constant2017multiword}. 
MWEs are valuable in that they contribute to enriching the expressiveness of a language, allowing diverse interpretations of their meaning according to the context in which they are located.
That is, the semantics of an MWE can be originated from either (i) the direct composition of the literal definitions of its constituents (i.e., compositional meaning) or (ii) its conventional usage in the language (i.e., idiomatic meaning).
For instance, given an expression called \textit{wet blanket}, its compositional meaning is `a piece of cloth soaked in liquid', whereas its idiomatic meaning is `a person who spoils the mood' (see Table \ref{table:table1}).

While MWEs function as an effective means of improving the abundance of a language, they are also one of the main obstacles that complicate natural language processing (NLP), from the perspective that an NLP model should be able to precisely identify their mode.
In addition, the current trend where the goal of most NLP models is chiefly focused on capturing compositionality raises the question of how properly to deal with idiomatic aspects of linguistic expressions \cite{garcia-etal-2021-assessing,garcia-etal-2021-probing,zeng-bhat-2021-idiomatic}.

\begin{table}[t!]
    \scriptsize
    \centering
    \begin{tabular}{l l l}
    \toprule
    \textbf{Category} & \textbf{Meaning} & \textbf{Example} \\ 
    \midrule
    \multirow{3}{*}{\shortstack[l]{Compositional \\ (Non-idiomatic)}} & \multirow{3}{*}{\shortstack[l]{A piece of cloth \\soaked in liquid.}} & And finally, the snow falls again, \\ 
    & & this time in a thick, \underline{\textit{wet blanket}} \\
    & & that encapsulates everything.\\
    \midrule
    \multirow{4}{*}{Idiomatic} & \multirow{4}{*}{\shortstack[l]{A person who \\spoils the mood.}} & When Marie brings him down to \\ 
    & & earth, it's not clear if she's being \\  
    & & a passive-aggressive \underline{\textit{wet blanket}}\\
    & & or if she might have a point. \\ 
    \bottomrule
    \end{tabular}
    \caption{Comparison between the compositional and idiomatic meanings of the expression \textit{wet blanket}.}
    \label{table:table1}
\end{table}

An intuitive solution for mitigating the aforementioned problem is an introduction of a sophisticated method designed to estimate the idiomaticity of a given expression, which enables the separate processing of the expression according to its category.
In a similar vein, we propose a series of techniques for better detecting the idiomaticity of a target MWE by actively exploiting its surrounding context in addition to considering the relationship between metaphors and the notion of idiomaticity.

\begin{figure*}[t!]
\begin{center}
\includegraphics[width=0.8\linewidth]{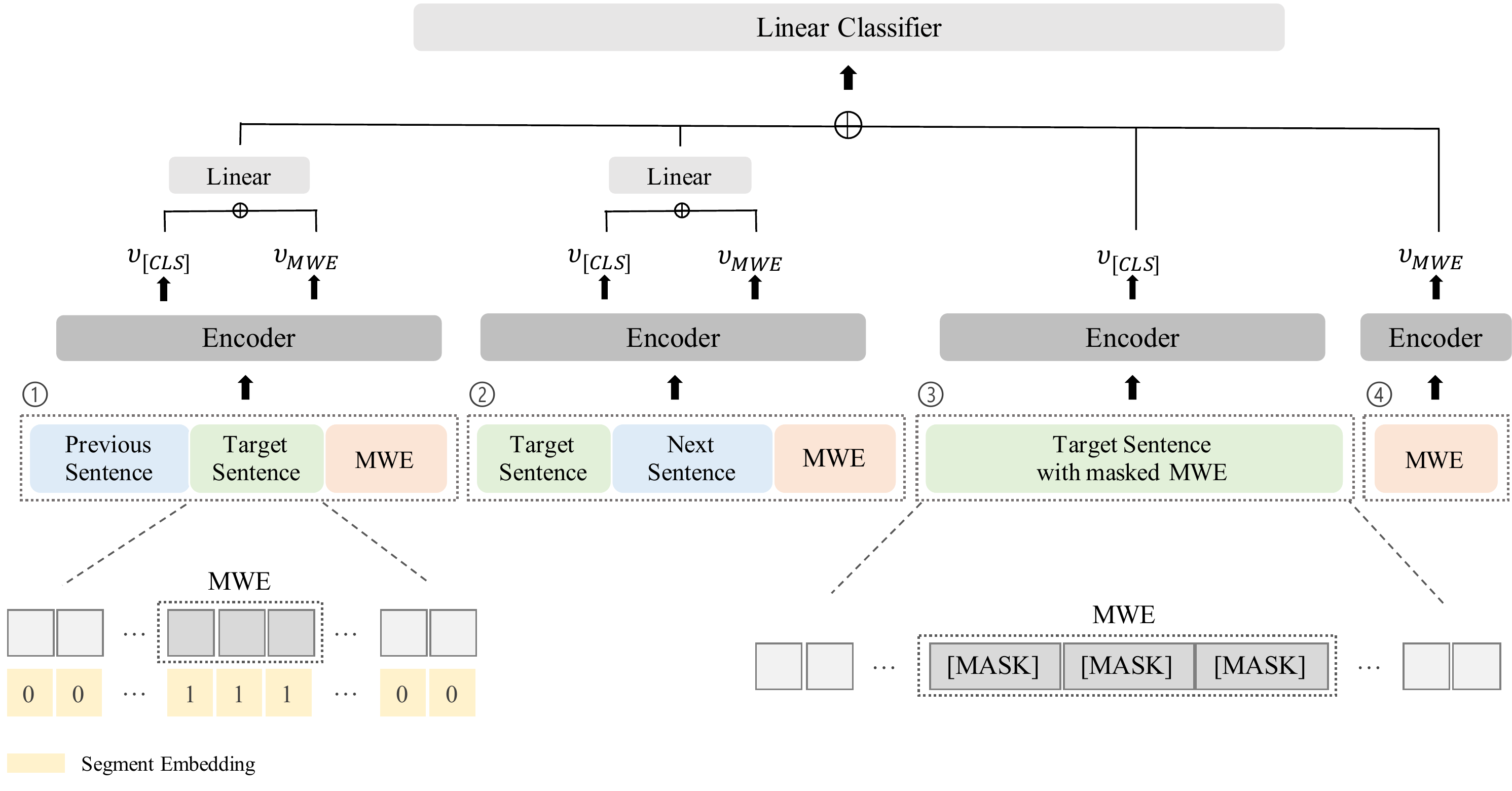}
\caption{Proposed framework. Two features on the left (\newtextcircled{1} and \newtextcircled{2}) are based on the surrounding context (Section \ref{subsec: features based on surrounding context}), while the remaining two (\newtextcircled{3} and \newtextcircled{4}) are originated from considering the inner-sentence context (Section \ref{subsec: features based on inner-sentence context}).} 
\label{fig:figure1}
\end{center}
\end{figure*}

Participating in SemEval-2022 Task 2, we focus on classifying two-noun compounds into idiomatic and non-idiomatic.
The task provides two configurations. 
In the zero-shot setting, a model's performance is evaluated on the set of sequences that include MWEs never appeared in the training phase. 
Meanwhile, in the one-shot setting, our model is exposed to a pair of instances per each MWE during training, one of which shows the idiomatic use of the MWE while the other is an example for the non-idiomatic case.\footnote{For more details on the task's specification, refer to the task description paper \cite{tayyarmadabushi-etal-2022-semeval}.}
We present a unified framework that can be used in both kinds, paying slightly more attention to the one-shot setting.
In extensive experiments, we show that most of our considerations lead to improvement in performance.
We also present discourse on the task specification to promote a fair comparison between related models.

\section{Related Work}

Idiomaticity detection has been widely studied in the literature \cite{reddy-etal-2011-empirical,liu2019generalized,garcia-etal-2021-assessing,garcia-etal-2021-probing,zeng-bhat-2021-idiomatic}. 
Above all, \citet{tayyar-madabushi-etal-2021-astitchinlanguagemodels-dataset} present a dataset that is the foundation of SemEval 2022 Task 2. 
This dataset consists of sentences that contain potential idiomatic MWEs with two surrounding sentences and annotations about the fine-grained set of meanings. 
The authors also evaluate a model's ability to detect idiomatic usage depending on whether context and MWE are included. 
They report that reflecting the context in the way of simply concatenating surrounding sentences is not generally helpful, and that adding the corresponding MWE at the end of the input sequence improves performance.
In the following sections, we re-examine their findings and present our own revision.

On the other hand, we investigate the viability of applying techniques for metaphor detection \cite{gao-etal-2018-neural,mao-etal-2019-end,lin-etal-2021-cate} into idiomaticity classification, inspired by the resemblance of the two tasks.
A metaphor is a form of figurative expression used to implicitly compare two things seemingly unrelated on the surface at the attribute level  \cite{baldwin2010multiword}. 
Not all metaphors have the property of idiomaticity, but some idioms rely on metaphorical composition.

In practice, \citet{choi-etal-2021-melbert} introduce two metaphor identification theories (Metaphor Identification Procedure (MIP; \citet{group2007mip}, \citet{steen2010method}) and Selectional Preference Violation (SPV; \citet{wilks1975preferential}) into their model to better capture metaphors, which we expect also might be helpful for the procedure of identifying idiomatic expressions. 
The basic ideas of MIP and SPV are that a metaphor can be identified when we discover the difference between its literal and contextual meaning, and that 
it can also be detected when its semantics is distinguishable from that of its context.
To realize the concepts, for MIP, \citet{choi-etal-2021-melbert} employ a target word's contextualized and isolated representations, while for SPV, they utilize the contextualized representations of the target word and the sentence including the word.
We adopt some of their ideas and customize them for our purpose, i.e., modeling features for idiomaticity detection.

\section{Proposed Method}

As a participant of SemEval-2022 Task 2, we propose a framework powered on four features devised to facilitate the detection of idiomatic expressions.
These features are computed by the same foundation model,\footnote{In this work, a `foundation model' refers to a Transformer \textit{encoder} pre-trained on large corpora, e.g., BERT and XLM-R.} but distinguished from each other by what is inserted into the model as input to compute the features.
A simple linear classifier is introduced on top of the concatenation of the four features to finally gauge the idiomaticity of an MWE in an input sequence.
Figure \ref{fig:figure1} presents the overall picture of our method.

\subsection{Features Based on Surrounding Context} \label{subsec: features based on surrounding context}

We first focus on the fact reported by \citet{tayyar-madabushi-etal-2021-astitchinlanguagemodels-dataset} that the \textit{surrounding context (a.k.a. inter-sentence context)}, which we define as sentences located right before and after a target sentence, is uninformative in predicting idiomaticity.
We hypothesize that this disappointing outcome is partly due to the way such context was exploited.

To be specific, given a sentence containing an MWE and its previous and following sentences, \citet{tayyar-madabushi-etal-2021-astitchinlanguagemodels-dataset} propose simply putting all the three together in order.
Despite its simplicity, this approach has an explicit drawback that a model should automatically learn how to distinguish the target sentence from its surrounding context.
Moreover, when combined with context without caution, the input sequence becomes much (approximately $3\times$) longer than its original form, which might cause a negative effect on performance by merely intensifying the complexity of the problem rather than providing additional information.

To alleviate the aforementioned problems, we suggest a new approach of combining a target sentence with its context. 
Concretely, we first concatenate the target sentence with its \newtextcircled{1} previous and \newtextcircled{2} next sentences \textit{respectively} (i.e., previous-target \& target-next), and then inject each chunk into our encoder to derive $v_{\text{[CLS]}}$ and $v_{\text{MWE}}$.
By doing so, we expect that the target sentence can be relatively more emphasized than its context, as the target sentence naturally appears twice while its context is exposed only once.
Plus, by dividing the whole sequence into two parts, it is anticipated that the encoder struggles less to extract useful information from the input.
Note that $v_{\text{[CLS]}}$ is the representation for the entire chunk, which is obtained by taking the [CLS] embedding from the last layer of the encoder, and that $v_{\text{MWE}}$ is the average of the representations of the subwords that constitute the target MWE.
Lastly, the final context-sensitive feature is computed by conducting a linear transformation of the concatenation of $v_{\text{[CLS]}}$ and $v_{\text{MWE}}$.

On the other hand, we propose two extra techniques in order to provide a clue on the location of MWEs.
While constructing token-level representations for our encoder, we employ trainable segment embeddings that draw the line between tokens for the target MWE and others.
Moreover, the target MWE is repeated at the end of each chunk, following \citet{tayyar-madabushi-etal-2021-astitchinlanguagemodels-dataset}.

\subsection{Features Based on Inner-Sentence Context} \label{subsec: features based on inner-sentence context}

Second, we consider adding features dedicated to more effectively leveraging the information embedded in the target sentence, regarding the MWE and its neighboring words as separate objects.
We import some ideas from prior work for metaphor detection \cite{mao-etal-2019-end,choi-etal-2021-melbert}, exploiting the conceptual relationship between metaphors and idiomatic expressions.

Initially, we assume that similar to Metaphor Identification Procedure (MIP), whose core idea is that a metaphoric word's semantics become distinct from its lexical meaning when it is contextualized, we consider an MWE as idiomatic when its static and contextualized embeddings are heterogeneous.
While the contextualized representation of the target MWE is already available from the features provided in Section \ref{subsec: features based on surrounding context}, we have not yet introduced the MWE's static representations.
To implement this, we again make use of the same encoder, however, only the MWE itself (removed from its context) is presented as input to the model. 
We call the output of this procedure the \newtextcircled{4} \textit{MWE-exclusive} representation, which becomes an ingredient for realizing the `idiomatic' version of MIP.
Note also that according to \citet{garcia-etal-2021-probing}, static models have been considered as competitive or even better to/than contextualized models for idiomaticity detection.
Therefore, we aim to reinforce our framework by employing both the options.

Meanwhile, \citet{choi-etal-2021-melbert} use Selectional Preference Violation (SPV) for metaphor detection, 
assuming that the semantics of a metaphoric word should be distinctive from that of its context.\footnote{This time, we limit the scope of the context as the sentence emcompassing a target expression (i.e., inner-sentence context), following \citet{choi-etal-2021-melbert}.}
We basically adopt their idea, but revise its implementation, arguing that their implementation might be somewhat defective.
In detail, \citet{choi-etal-2021-melbert} compute the embeddings of a target and its context exactly as we do when computing $v_{\text{[CLS]}}$ and $v_{\text{MWE}}$ in Section \ref{subsec: features based on surrounding context}.
However, it is highly probable that $v_{\text{[CLS]}}$ and $v_{\text{MWE}}$ contain similar information as they are intertwined with each other by the attention mechanism, which is undesirable when estimating separate semantics of the target and context.
We thus introduce the \newtextcircled{3} \textit{context-exclusive} representation of the target sentence by providing our encoder with a variant of the sentence where the target MWE is masked.
When combined with the features from the previous section, we expect that the inner-sentence context independent from the target MWE at all can be useful for applying the concept of SPV into idiomaticity detection.

\section{Experiments}

\subsection{Experimental Setup}










For all experiments, we present five instances per each model with the corresponding random seeds (42, 360, 2578, 5925, 9463). 
Each instance is trained for 10 epochs, and its best checkpoint which shows the top performance on the development set in terms of the macro F1-score is chosen for the inference of the test set. 
We use a max sequence length of 300, a learning rate of 3e-5 for the AdamW \cite{loshchilov2019decoupled} optimizer, and a batch size of 16 for the training set and 8 for the validation and test sets. The vectors of each representation ($v_{\text{[CLS]}}$ and $v_{\text{MWE}}$) have 768 dimensions respectively and the learning rate is scheduled to linearly decrease after the second epoch.
We leverage XLM-R(-base) \cite{conneau-etal-2020-unsupervised} as our foundation model. 

\subsection{Main Results}

\begin{table}[t!]
    \small
    \centering
    \setlength{\tabcolsep}{0.25em}
    \begin{tabular}{l c c c c}
    \toprule
    Model / Lang. & English & Portuguese & Galician & Overall \\ 
    \midrule
    \textit{Zero-shot setting} \\
    Baseline (BERT) & 70.70 & 68.03 & 50.65 & 65.40 \\
    Baseline (XLM-R) & 72.29 & 65.68 & 46.16 & 63.21 \\
    Ours (submitted) & \textbf{76.42} & \textbf{72.82} & \textbf{62.92} & \textbf{72.27} \\ 
    \midrule
    \textit{One-shot setting} \\
    Baseline (BERT) & 88.62 & 86.37 & 81.62 & 86.46 \\
    Baseline (XLM-R) & 88.45 & 85.03 & 84.02 & 86.56 \\
    Ours (submitted) & 91.59 & 84.57 & 82.87 & 87.50 \\ 
    Ours (post-eval) & \textbf{92.29} & \textbf{88.05} & \textbf{87.10} & \textbf{89.96} \\ 
    \bottomrule
    \end{tabular}
    \caption{Main results on the test set. Numbers are from the best configuration (random seed) of each model.}
    \label{table:table2}
\end{table}

We compare the results of our method (submitted) against those of the baseline offered by the task organizers \cite{tayyarmadabushi-etal-2022-semeval}.
Although the original baseline is powered on Multilingual BERT(-base), for a comparison, we also report the performance of the baseline equipped with XLM-R(-base).
Evaluation is conducted on the test set, and each model's performance is reported according to the language on which it is tested (English, Portuguese, and Galician) and the setting it is trained (zero- and one-shot).

From Table \ref{table:table2}, we can see that both in the zero- and one-shot settings, our model largely outperforms baselines. 
Notice that in the zero-shot setting, our model outperforms the baseline powered on the same foundation model (XLM-R) by more than 16\% in Galician. 
Considering that Galician was not included in the training data, this result confirms that our model is more generalizable than the baselines from the perspective of input language.

\begin{figure}[t!]
\begin{center}
\includegraphics[width=\linewidth]{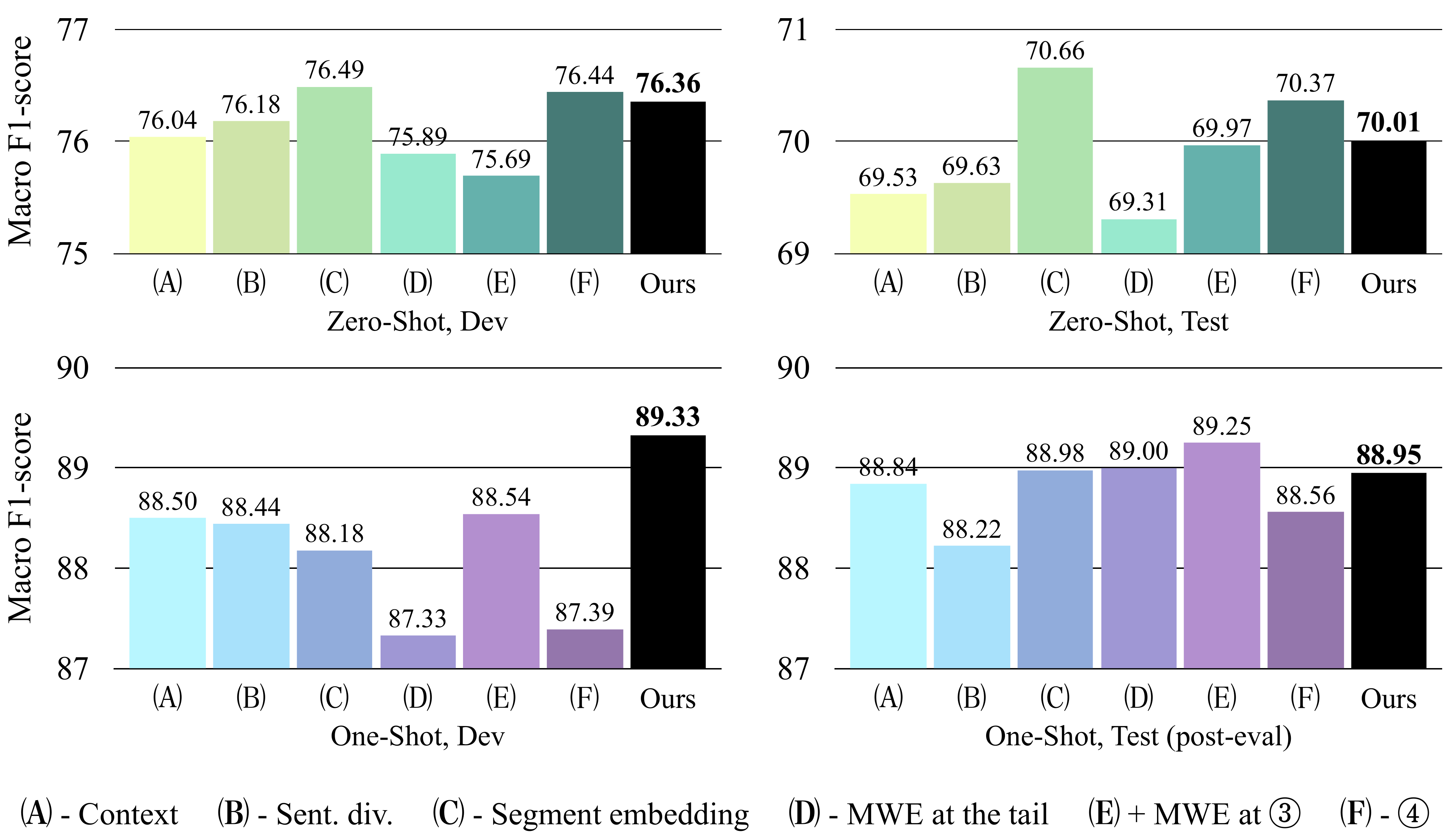}
\caption{Ablation study.} 
\label{fig:figure2}
\end{center}
\end{figure}

\subsection{Ablation Study}

We perform an ablation study to confirm whether the elements of our framework are significant. 
Note that all the results used for comparison are the average of the scores of different instances with five random seeds.
Overall, when tested on the validation set, the final version of our approach succeed in outperforming most of the variations, especially in the one-shot setting where our decisions for selecting the final model were made. 
We present more detailed analysis in the following.

First, we compare our method with the variation (A) which uses only target sentences without surrounding context and the variation (B) which reflects the context by concatenating three sentences. 
\citet{tayyar-madabushi-etal-2021-astitchinlanguagemodels-dataset}, where the authors employ the variation (B), previously reported that it is not helpful for idiomaticity detection to consider the surrounding context of a target MWE.
However, as shown in Figure \ref{fig:figure2}, we find that taking the context into account following our approach (i.e., separating the context into two chunks) is in fact advantageous in all experimental settings. 
Furthermore, we observe that the deviation of the scores of our method is much smaller and more stable than that of not considering context. 
This implies that if there exists a data instance not having much information available from its target sentence, the surrounding context of the target sentence can complements the lack of information.

Contrary to our expectation, it is shown that our method is not always better than the three variations (C), (D) and (E).
The variation (C) removes segment embeddings, the variation (D) stops the repetition of MWEs at the tails of \newtextcircled{1} and \newtextcircled{2}, and in the variation (E)
the target MWE is recovered (not masked) in the computation of  `\textit{context-exclusive}' representation (\newtextcircled{3}).
We leave a detailed examination regarding these as a follow-up study.

Lastly, it turns out that the variation (F) which removes the `\textit{MWE-exclusive}' representation (\newtextcircled{4}), is more favored in the zero-shot setting. 
Unlike the one-shot setting, where a pair of positive and negative examples for a particular MWEs can be provided, the zero-shot setting requires the evaluation of MWEs not presented in the training set, which is a more harsher condition for idiomaticity detection models.
Therefore, we conjecture that static representations for the MWEs unseen during training become a little bit noisy in the zero-shot setting, failing to function following our intention.

\section{Discussion}

\subsection{Issue on Validation Set in One-shot Setting}

In the one-shot setting, we expect that a pair of data instances (one for idiomatic and the other for non-idiomatic) per every MWE in the test set should be provided to the model we train.
Likewise, if one wants to confirm that the validation set is rigorous enough to be a substitute for the test set in the procedure of selecting hyperparameters, every MWE in the validation set should have two corresponding instances in the training set.
During the competition for SemEval 2022 Task 2, we have discovered that the necessary condition holds for the validation set in the practice phase, while \textbf{it does not hold in the evaluation phase}.
In other words, the training set provided in the practice phase incorporated data instances that correspond to MWEs in the validation set.
However, as the training set has been substituted with a new version, a problem has arisen where MWEs in the newly released training set do not match with those in the validation set.
We conjecture that this discrepancy prevents one's optimal actions in choosing the best models.

To prove our hypothesis, we test a variant whose performance on the validation set is not optimal, but has the potential of working well when evaluated on the test set.
Specifically, we replicate our experiments, but do not choose the best instance based on validation performance. 
Instead, we simply choose the model instance trained until 9 epochs and compare it to baselines.
As shown in Table \ref{table:table2}, we find that the instance chosen based on the validation set (i.e., ours (submitted)) is worse than the randomly selected one (i.e., ours (post-eval)), implying that the inappropriateness of the validation set in the evaluation phase might hinder correct comparisons between submitted models. 

\begin{table}[t!]
    \small
    \centering
    \begin{tabular}{l c c}
    \toprule
    Form of MWEs & Validation & Test\\ 
    \midrule
    \textit{Zero-shot setting} \\
    Original form & 76.34 & \textbf{71.72}\\
    Inflectional form & \textbf{76.36} & 70.01\\
    \midrule
    \textit{One-shot setting} \\
    Original form & 88.14 & \textbf{89.80}\\
    Inflectional form & \textbf{89.33} & 88.95\\
    \bottomrule
    \end{tabular}
    \caption{Performance gap with form changes in MWEs.}
    \label{table:table3}
\end{table}

\subsection{Impact of Form Changes in MWEs}

When MWEs are repeated at the end of input sequences in \newtextcircled{1} and \newtextcircled{2} and embedded solely in \newtextcircled{4} in our implementation, we copy them from target sentences so that we can preserve their inflectional form appearing in the sentences, rather than adopting their original form.
To confirm the effectiveness of this approach, we conduct an experiment where we replace the MWEs with their original form. 

From Table \ref{table:table3}, we observe that unlike the case on the validation set where applying inflectional form is always helpful, it turns out that when evaluated on the test set, employing inflectional form is not beneficial for performance improvement, contrary to our expectation.
The idea of having utilized the inflectional form of MWEs is from our conjecture that compositional and static representations of MWEs should be computed from the same form for a fair comparison between them.
However, it seems that it is more effective to provide a model with a MWE's original form in addition to its inflectional form such that the model can extract more information from the both sides.
We leave thorough analysis on this phenomenon as future work.

\section{Conclusion}

In this work, we investigate the method of implementing better idiomaticity detection models by considering different levels of contextualization.
We propose four features grounded on the surrounding and inner-sentence context of a target MWE, showing that these features are effective in improving performance.
Moreover, we present a discussion on the issue related to the validation set in the one-shot setting and the impact of the form of MWEs.
As future work, we plan to develop a method of designing the interaction between related features in a more sophisticated fashion, instead of simply concatenating them.

\section*{Acknowledgements}

This work was supported by Institute of Information \& communications Technology Planning \& Evaluation (IITP) grant funded by the Korea government(MSIT) (No.2020-0-01373, Artificial Intelligence Graduate School Program (Hanyang University)).

\bibliography{references}




\end{document}